\documentclass[letterpaper, 10 pt, conference]{ieeeconf}  

\IEEEoverridecommandlockouts                              

\usepackage{etoolbox}
\makeatletter
\patchcmd{\@makecaption}
  {\scshape}
  {}
  {}
  {}
\makeatletter
\patchcmd{\@makecaption}
  {\\}
  {.\ }
  {}
  {}
\makeatother
\usepackage[format=plain]{caption}
\usepackage{graphics} 
\usepackage{epsfig} 
\usepackage{mathptmx} 
\usepackage{times} 
\usepackage{amsmath} 
\usepackage{amssymb}  
\usepackage{algpseudocode}
\usepackage{algorithm}
\usepackage{url}

\usepackage{graphicx}
\title{\LARGE \bf
Parameter Sharing Reinforcement Learning Architecture for Multi Agent Driving Behaviors
}
\author{Meha Kaushik$^{1}$, Phaniteja S$^{2}$ and K. Madhava Krishna$^{3}$ 
\thanks{All authors are affiliated with The Robotics Research Center, International Institute of Information Technology, Hyderabad.
       {\tt\small meha.kaushik@research.iiit.ac.in}$^{1}$ 
       {\tt\small singamaneniphani.teja@research.iiit.ac.in}$^{2}$
       {\tt\small mkrishna@iiit.ac.in}$^{3}$
       }
}
\begin{document}
\maketitle
\thispagestyle{empty}
\pagestyle{empty}

\begin{abstract}

Multi-agent learning provides a potential framework for learning and simulating traffic behaviors. This paper proposes a novel architecture to learn multiple driving behaviors in a traffic scenario. The proposed architecture can learn multiple behaviors independently as well as simultaneously. We take advantage of the homogeneity of agents and learn in a parameter sharing paradigm. To further speed up the training process asynchronous updates are employed into the architecture. While learning different behaviors simultaneously, the given framework was also able to learn cooperation between the agents, without any explicit communication. We applied this framework to learn two important behaviors in driving: 1) Lane-Keeping and 2) Over-Taking. Results indicate faster convergence and learning of a more generic behavior, that is scalable to any number of agents. When compared the results with existing approaches, our results indicate equal and even better performance in some cases.

\end{abstract}
\section{Introduction and Related Work }
Reinforcement Learning (RL) algorithms when trained with the correct reward functions and favorable for learning, training conditions, have shown surprisingly impressive results. Some popular examples being: mastering the Go Game \cite{silver2016mastering}, playing Atari Games \cite{mnih2013playing} and the very recent, defeating the world's top professionals in DOTA \cite{OpenAI_dota}. RL has shown promising results in learning driving behaviors for single agents \cite{xia2016control}, \cite{sallab2017deep}, \cite{meha2018iv}, \cite{sharifzadeh2016learning}, \cite{loiacono2010learning}, \cite{tai2017virtual}. Inspired from the same, this work focuses on behavioral based learning in multi agent settings.


A lot of prior work exists for multi agent systems \cite{busoniu2006multi,bloembergen2015evolutionary,ono1997modular}. The major paradigms include frameworks which use inter agent communication \cite{sukhbaatar2016learning,foerster2016learning}, and the ones which learn in a decentralized manner \cite{lauer2000algorithm} and ones which learn in a centralized manner \cite{moradi2016centralized}, there also exist frameworks where training is centralized but testing is decentralized \cite{foerster2017counterfactual,lowe2017multi}. Centralized learning refers to learning actions jointly for all the agents. The input to the algorithm is the observation and action of all the agents, which results in a major disadvantage: the exponential increase in state space with the number of agents. Secondly, centralized approaches are centralized not only during training but during testing as well, resulting in higher resource requirements for actual deployments. Unlike centralized approaches, in a concurrent learning setting, multiple agents in the same environment learn independently. Each one of them will have their own networks, policies, observations and actions. This is equivalent of learning multiple single agent learnings in a same environment. The disadvantage of this approach is the huge number of parameters and  that no advantage is drawn from the fact that agents are learning together. Lastly, each agent is learning independently, hence the environment is non-stationary which can lead to instability.

When similar agents are learning similar behaviors, their parameters can be shared to enhance the speed of learning and to decrease the complexity and resource utilization of the algorithm. This concept of parameter sharing was first introduced by Tan et.al in \cite{hu1998multiagent}. Authors showed that if cooperation is done intelligently, each agent can benefit from other agents' instantaneous information, episodic experience, and learned knowledge. Sharing learned policies and episodes between agents can speed up the whole learning. Policies can be shared between homogeneous agents only, and if episodes can be interpreted, heterogeneous agents can also benefit from sharing episodes. 

Chu et.al have shown Parameter Sharing in special cases in \cite{chu2017parameter}.
Recently, Gupta et.al \cite{gupta2017cooperative} introduced Parameter Sharing extensions of three popular RL algorithms: Deep Q-Network (DQN) \cite{mnih2015human}, Asyncronous Advantage Actor-Critic (A3C) \cite{mnih2016asynchronous} and Trust Region Policy Optimization (TRPO) \cite{schulman2015trust}. Their results indicate a scalable cooperative reinforcement learning algorithm, Parameter-Sharing TRPO and also show that Policy Gradient methods outperform temporal-difference and actor-critic methods. Inspired by their success, we have developed a Parameter Sharing Deep Deterministic Policy Gradients (DDPG) \cite{lillicrap2015continuous} architecture, where the agents share their Actor and Critic Networks. 

The proposed architecture was applied to traffic agent behaviors where multiple homogeneous agents are learning similar behavior, which is trained by asynchronous and cumulative efforts of all agents. The primary motivation behind the proposed work is to develop a method which can be used to generate behavior based traffic in simulators. With the rising interests in autonomous driving research, simulated environments provide a fast and risk-free method to develop and test the algorithms. To the best of our knowledge, this is the first work that targets behavior based multi agent learning using Deep RL. Further, results indicate faster trainings, scalable learning, which can be  tested with varied number of agents, independent from the number of agents trained earlier. Importantly, the architecture is able to learn multiple behaviors simultaneously using single Actor-Critic Networks.  
 
The rest of the paper is organized as follows, Section II explains the architectural details of our proposed approach and Section \ref{implementation} contains the details of implementation specific to behavior learning for driving agents and finally section \ref{results} shows the results of various experiments from our approach.
\section{ Proposed Architecture} \label{architecture_section}
Multi agent learning is a challenging task because of the dynamic nature of the environment. Each agent explores the environment in an attempt to learn a policy which increases the complexity of learning for other agents. Above everything, learning for multiple agents involve high number of parameters and resource requirements, which further limits the performance of the algorithms. In this section we propose and explain an architecture that addresses these problems using the concept of parameter sharing. The proposed architecture is based on Deep Deterministic Policy Gradients (DDPG) \cite{lillicrap2015continuous}, one of the first RL algorithms which targeted problem solving in continuous spaces. It has shown promising results in wide ranged domains: Humanoids \cite{phaniteja2017deep}, controlling a bicycle \cite{chung2017controlling} and the most relevant here, driving on tracks \cite{sallab2017deep} and overtaking behavior in presence of other cars \cite{meha2018iv}. 

\begin{figure}[!h]
\begin{center}
\includegraphics[width=1\columnwidth]{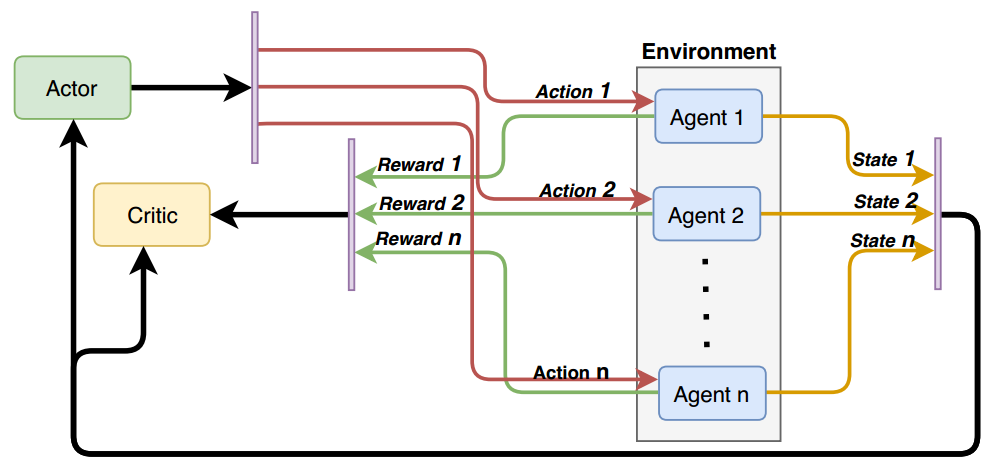}
\caption{\small{Parameter Shared DDPG (PS-DDPG). The purple bars in the figures lets only one signal to pass at one point of time, depending on which agent is selected.}}
	\label{architecture}
\end{center}
\end{figure}

The proposed architecture is shown in Fig \ref{architecture}. As shown in the figure, both Actor and Critic Network are shared between all the agents. Apart from these, we also maintain a shared Replay Buffer, which stores the experiences from all the agents. Each agent has its own copy of its state information, its observations from the environment, the actions it takes and its corresponding rewards. For a given agent this information is not known to any other agent. However, the data stored in Replay Buffer is not distinguishable and hence each agent gets benefits from the experiences of all agents. Finally, actor and critic networks are updated asynchronously by each agent at each step. The update equations are given as follows:

\begin{itemize}
\item The Critic Network learns by minimising the loss between target $y$ and the current Q value:
\begin{equation}
\begin{split}
& L = \frac{1}{N}\displaystyle\sum_{t}( y_{i, t} - Q(s_{i, t},a_{i, t}) )^{2} \\
& y_{i , t} = (r_{i , t} + \gamma Q_T(s_{i,t+1},\mu_T(s_{i,t+1})))
\end{split}
\label{eq:critic}
\end{equation}
where $r_{i,t}$ is the reward for $i^{th}$ agent at the $t^{th}$ timestep, $Q_T(s_{i,t+1},\mu_T(s_{i,t+1}))$ is the target Q value for the state-action pair $(s_{i,t+1},\mu_T(s_{i,t+1}))$ where $\mu_T(s_i,{t+1})$ is obtained from the target actor network, $Q(s_{i,t},a_{i,t})$ is the Q value from the learned network, $N$ is the batch-size and $\gamma$ is the discount factor.

\item The Actor Network weights are updated as: 
\begin{equation}
\begin{split}
& \theta\mu = \theta\mu + \alpha \nabla_{\theta\mu}J\\
& \nabla_{\theta\mu}J \approx \frac{1}{N}\displaystyle\sum_{t} \nabla_{a}Q(s,a)|_{s=s_{i,t},a=\mu(s_{i,t})}
\nabla_{\theta^\mu}\mu(s)|_{s=s_{i,t}}
\end{split}
\label{eq:actor}
\end{equation}
where $N$ is the batch-size, $\theta^{Q}$ are the critic network parameters and $\theta^{\mu}$ are the actor network parameters, $\alpha$ is the learning rate. The rest of the terms have the same meaning as those in Eq. \ref{eq:critic}.
\end{itemize}
The reward function has to be represented using one standard function for all the agents, independent of the behavior they learn. Additionally, the reward function should only contain variables which can be derived from the state information and the observation of the agent. This condition, makes sure that the experiences in the replay buffer could be generalized to all agents.

The proposed setting is highly advantageous over multiple DDPG \footnote{Learning for multiple agents in a same environment using independent DDPGs for each of them} setting. Firstly, in each step, every agent is updating the networks, hence the speed of training is increased by $n$ times, where $n$ is the total number of agents learning. In a multiple DDPG setting, since each agent maintains a separate Actor and Critic, only one update is possible for the corresponding networks in each step. Hence, it takes longer time to converge. Moreover, because of multiple Actor and Critic Networks, very large number of parameters are present in the architecture.

Secondly, the agents in the proposed architecture use a shared replay buffer. Sharing the replay buffer increases the diversity of experience for all the agents. This way, the learned behavior of one agent does not depend only on the experiences it sees, rather on the experiences of all the agents which are getting trained. Sharing is possible, since the agents are homogeneous in their properties. Unlike this setting, Multiple DDPG do not have a shared replay buffer and depends only on the agent’s individual experiences even when the agents are homogeneous. This is another drawback in this setting, even though the agents are learning in multi agent setting, they do not make use of it for faster learning.

\section{Implementation Details} \label{implementation}

We use a modified version of TORCS called Gym-TORCS \cite{GymTORCS} which supports development of  RL algorithms. The agent car used is "scr\_server". We use a NVIDIA GeForce GTX 1080 GPU for training. 

For individual behavior learning (either Lanekeeping or Overtaking) , the state vector is a 65 sized array consisting of the following sensor data:
\begin{enumerate}
\item \textbf{Angle} between the car and the axis of the track. 
\item \textbf{Track Information:} Readings from 19 sensors with a 200m range, present at every $10^\circ$ on the front half of the car. They return the distance to the track edge.
\item \textbf{Track Position:} Distance between the car and the axis of the track, normalized with respect to the track width. 
\item \textbf{SpeedX, SpeedY, SpeedZ}
\item \textbf{Wheel Spin Velocity} of each of the 4 wheels.
\item \textbf{Rotations per minute} of the car engine 
\item \textbf{Opponent information:} Array of 36 sensor values, each corresponding to the distance of the nearest opponent in the range of 200 meters, located at a difference of $10^\circ$, spanning the complete car. 
\end{enumerate}
Further details about each of these sensor readings can be found in \cite{loiacono2013simulated}. The Action Vector consists of continuous values of \textbf{steer} (-1,1), \textbf{acceleration} (0,1) and \textbf{brake} (0,1).
\subsection{Reward functions for the Behaviors learned}
For all of our experiments, we have used two main reward functions. Both of these have been inspired from the work done in \cite{meha2018iv}.
\subsubsection{Lanekeeping}
Lanekeeping is a behavior when the agent drives straight on the road and it is motivated by the distance it moves along the lane in each step. The Reward function to learn this behavior is given by: 
\begin{equation}
\label{eq:lane}
R_{Lanekeeping} = v_{x}(cos\theta - sin\theta) 
\end{equation}
where $v_{x}$ denotes the longitudinal velocity of the car, $\theta$ denotes the angle between the car and the track axis. We give a positive reward when the car moves forward along the track axis, given by $v_{x}cos\theta$, and negative reward when it moves laterally, i.e. perpendicular to the track axis, given by $-v_{x}sin\theta$. 
The above function can standalone handle the negative impact conditions like collisions, off track drifting, since on colliding with walls or other agents, ego vehicle's velocity will be decreased and hence the above term. The decrease whether significant or not, the velocity has high probability of remaining positive. The learning algorithm would take high number of episodes to understand that collisions are bad. To increase the learning, we introduce extra reward conditions for such not required cases. 
\begin{table}[h]
\caption{Extra Rewarding Conditions}
\label{discrete}
\begin{center}
\begin{tabular}{|p{2cm}|p{2.5cm}|}
\hline
Condition & Reward\\
\hline
Collision & $-1000$ \\
Off track drifting & $-1000$\\
No Progress & $-500$\\
\hline
\end{tabular}
\end{center}
\end{table}
\subsubsection{Overtaking}
We used the reward of overtaking in \cite{meha2018iv}. The reward function in this case is given by:
\vspace{-0.2em}
\begin{equation}
\label{eq:ovetaking}
R_{overtaking} = R_{Lanekeeping} + 100*(n - racePos)
\end{equation}
Here $n$ denotes total number of cars in a given episode and $racePos$ denotes the position of car in the race, which is obtained from the simulator. The extra reward conditions are same as in table \ref{discrete}. 
\subsubsection{Multi-Behavior Learning}
By multi-behavior, we imply learning multiple behaviors simultaneously using one single instance of the architecture.
For Multi-Behavior learning the type of agents have to be distinguished somehow. For the same, we gave them a ids. Agents which had to learn the overtaking behavior were given id as 1 and the lanekeeping agents were given the id as 0.  The state vector was modified from 65 to 66 space, because of the addition of id. Reward function should be a single equation using the terms derivable from observation or state vector of the agent. Following this,
\begin{equation}
\label{eq:ovetaking2}
R_{multi} = R_{Lanekeeping} + id*(100*(n - racePos))
\end{equation}
For a given training, $n$ is the total number of agents present in the simulator, which is a constant term, hence satisfies the requirement of permissible variables in the reward function. Next, $racePos$ is a term TORCS provides as observation for each agent, hence this variable also satisfies the requirement.
\section{RESULTS} \label{results}
\subsection{Lanekeeping Behavior}
\begin{itemize}
\item Figure \ref{lanekeeping} depict the result of our architecture. We learned lanekeeping behavior for 6 agents and tested it for number as high as 20. The agents moved harmoniously with minimal collisions and followed the lane, staying in the middle maximal times.
\begin{figure}[H]
\includegraphics[width=\linewidth]{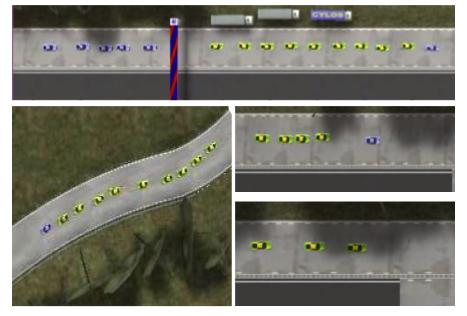}
\caption{\small{Various instances of lanekeeping results from TORCS, the blue and green cars, all are the trained agents. The training was done on 6 agents, the results are shown on 3, 5 10 and 15 agents.}}
	\label{lanekeeping}
\end{figure}

\item Table \ref{lane-learning} shows how the learning has evolved with training episodes. At episode 0, the agents starts to train, hence the reward is 0 while testing. Till 300 episode of training the agents have learned to drive on lane, with collisions only around 11\% times, after 600 episodes of training the average sum of reward of the complete system has improvised, and  collisions are approximately same. From 300 to 600,  the training results have almost saturated. The reward indicates sum of reward of all agents over an episode. While training there were 6 learning agents in the environment. 
\begin{table}[H]
\begin{center}
\begin{tabular}{|c|c|c|c|}
\hline
\begin{tabular}[c]{@{}c@{}}No. of\\ Training\\ Episodes\end{tabular} & \begin{tabular}[c]{@{}c@{}}Sum of \\ Reward \\ of all\\ agents\end{tabular} & \begin{tabular}[c]{@{}c@{}}\%colliding\\ steps\\ in the \\ system\end{tabular} & Observations                                                       \\ \hline
0                                                                    & 0                                                                           & 0                                                                              & Nothing learned                                                    \\ \hline
300                                                                  & 47476                                                                       & 10.89                                                                          & \begin{tabular}[c]{@{}c@{}}Learns to drive on \\ lane\end{tabular} \\ \hline
600                                                                  & 50180                                                                       & 11.4                                                                           & More stable driving                                                \\ \hline
\end{tabular}
\end{center}
\caption{\small{The percentage increase in reward from 300 to 600 episodes, is 5\%, indicating over time reward value starts saturating. All the values are averaged over 20 episodes. The sum of reward of all agents is calculated episode wise, averaged over 20 episodes. \%colliding steps indicate the times when any one or more agents were experiencing collisions}}
\label{lane-learning}
\end{table}

\begin{table}[!ht]
\begin{center}
\begin{tabular}{|c|c|}
\hline
\begin{tabular}[c]{@{}c@{}}Number  \\ of Agents\end{tabular} & \begin{tabular}[c]{@{}c@{}}Average total\\ Reward/Progress\\ per agent\end{tabular} \\ \hline
3                                                            & 8165.8                                                                              \\ \hline
5                                                            & 7903.8                                                                              \\ \hline
7                                                            & 7918.6                                                                              \\ \hline
10                                                           & 7510.3                                                                              \\ \hline
12                                                           & 7570.8                                                                              \\ \hline
15                                                           & 7678                                                                                \\ \hline
\end{tabular}
\caption{\small{Total reward of each agent in an episode, averaged over 600 episodes, for lanekeeping behavior. We observe as the number of agents increase the total reward for each agent is approximately same, this indicates that with increasing number of agents, the per agent reward is not getting affected, implying behavior and performance of agents is not affected. This demonstrates both scalability and stability of our approach.}}
\label{lane-testing}
\end{center}
\end{table}
\item Lastly, we compare PS-DDPG with regular DDPG, trained in a single agent environment and multiple DDPG agents trained in together.
\begin{figure}[H]
\centering
\includegraphics[width=\columnwidth]{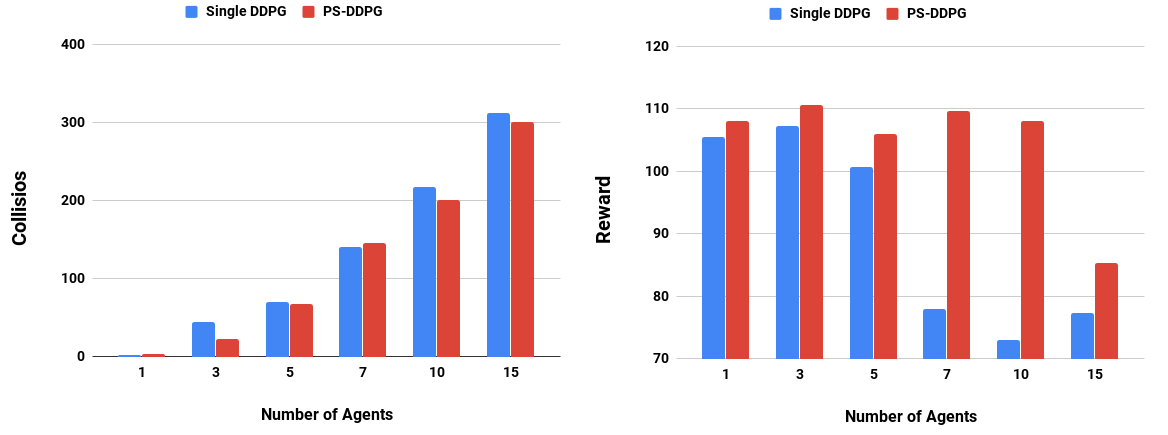}
\caption{\small{Comparing results of PS-DDPG and DDPG. The x-axis in both plots is number of agents. The plot on the left shows the number of collisions as the number of agents increase and the right one shows the average reward in the same settings. In the figure, blue bars correspond to single DDPG and red bars to PS-DDPG.  The results are produced after 300 episodes of training using PS-DDPG with 8 agents and after 2k episodes of training of single DDPG. The evaluation is done by varying the number of agents. In terms of total number  of collisions the complete system sees, both of the approaches perform similar. As one can expect, with the increase in number of agents, total collisions increase for both approaches. In the second graph, we compare the total reward per agent, per time step, averaged over 20 episodes. With the increase in number of agents, this value decreases for DDPG, indicating, single agent DDPG cannot support stable scalability. PS-DDPG, on the contrary, has a relatively stable value of rewards. }}
\label{comp_lane}
\end{figure}
\end{itemize}
In multi-DDPG setting, when 6 agents were trained simultaneously, only the 3 agents in the front were able to learn the behaviour. The last 3 agents could not learn to drive and got stuck at local minima. During training they collided with the front cars and gained negative rewards, henceforth they learned not to move forward at all. In this setting, the average reward per agent is highest in case of single agent with 79.3257 and declined as more agents are introduced into the setting. In case of 7 agents, the average reward obtained in 42.6297. However, in PS-DDPG setting, the average reward remained almost constant, as evident from the Fig. \ref{comp_lane}

The number of training episodes required for DDPG were 2000, for PS-DDPG were 300 and for Multiple DDPGs together were 3000. 
\subsection{Overtaking behavior}
\begin{figure*}[h]
\begin{center}
\includegraphics[width=1\textwidth, height=6cm]{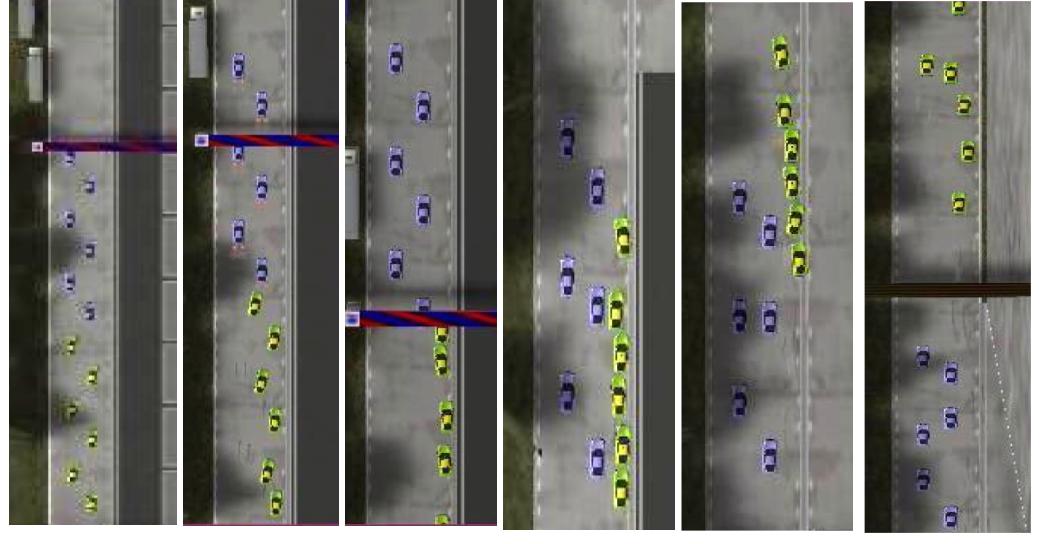}
\caption{\small{In the above figure, the yellow cars have learned overtaking behavior. The blue cars are passive opponent agents. All the yellow cars start from behind the blue cars, they move to left side of road, overtake the opponent cars and comeback to middle of the road.}}
	\label{overtaking}
\end{center}
\end{figure*}
Similar to the curriculum learning approach followed in for overtaking behaviors in \cite{meha2018iv}, we initialized our overtaking learning with weights from lanekeeping learning. Curriculum learning not only helps in learning the behavior but also reduces the training time. 
\begin{itemize}
\item Figure \ref{overtaking} shows results of our approach for overtaking behavior. Our learned agents, align themselves towards right end, overtake the opponent agents and scatter back on the road. 
\item Evaluation of how the training progresses is done in table \ref{overtake-training}. Overtaking behavior is learned in a curriculum fashion, it starts with initialization of lanekeeping weights. Hence, the total system reward is not zero, unlike the lanekeeping case. Overtaking, being a more complex behavior, is effectively learned in 600 episodes, unlike lanekeeping which was learned in 300 episodes only.
\begin{figure*}[h]
\begin{center}
\includegraphics[width=2\columnwidth]{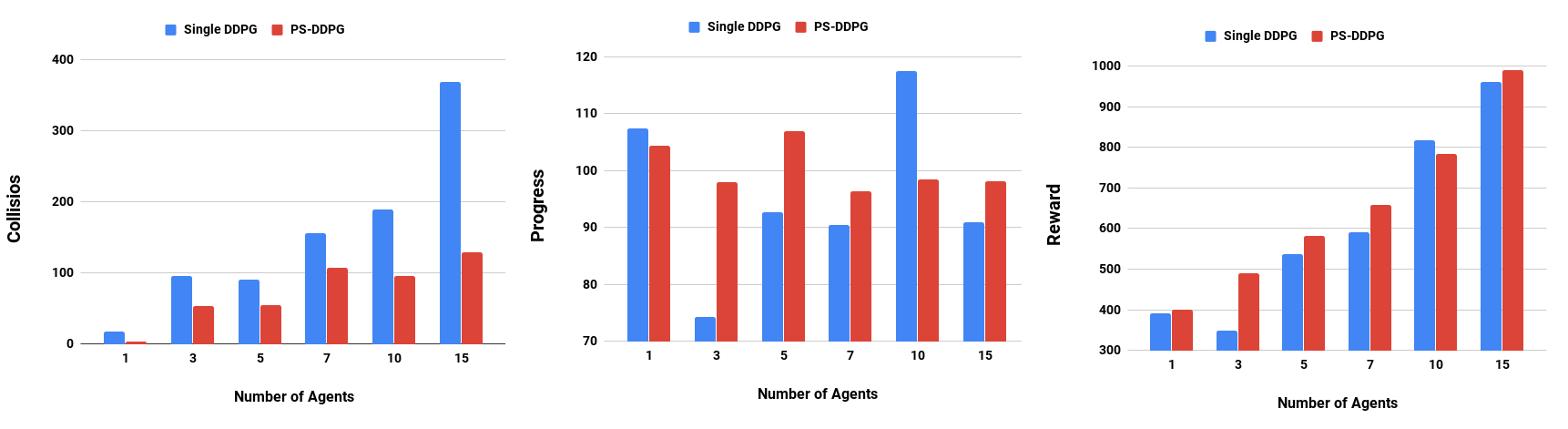}
\label{comp_overtake}
\caption{\small{The blue bars in all the plots correspond to single DDPG setting while the red ones correspond to PS-DDPG. The figure shows number of collisions, progress and average reward against increasing number of agents. The values Reward and Progress are calculated for each time step, averaged over 20 episodes and number of agents indicate the agents which are using the trained network for their control. We observe single agent DDPG performs better in case of 1 agent, but as the number of agents increase, the performance in terms of collisions and progress decrease drastically for DDPG. Contrarily, PS-DDPG performs better, this justifies not only scalability but generality of performance i.e. PS-DDPG can perform across wide variety of situations. The reason of the same can be attributed to the multi agent setting in training which helps it to explore better and see diverse set of experiences. The reward values are linearly increasing because of the reason mentioned in \ref{overtake-testing}. Overall, the above graphs indicate better performance for PS-DDPG then DDPG.} }
 \label{overtake-compare}
 \end{center}
\end{figure*}
\begin{table}[H]
\begin{center}
\begin{tabular}{|c|c|c|c|c|}
\hline
\begin{tabular}[c]{@{}c@{}}No. of\\ Training\\ Episodes\end{tabular} & \begin{tabular}[c]{@{}c@{}}Sum of \\ Reward \\ of all\\ agents\end{tabular} & \begin{tabular}[c]{@{}c@{}}Sum of \\ Progress\\ of all\\ agents\end{tabular} & \begin{tabular}[c]{@{}c@{}}\%colliding\\ steps\\ in the \\ system\end{tabular} & Observations                                                                                                        \\ \hline
0                                                                    & 241420                                                                      & 43516                                                                        & 21.6                                                                           & \begin{tabular}[c]{@{}c@{}}Follows lane-\\ keeping behavior\end{tabular}                                            \\ \hline
300                                                                  & 228160                                                                      & 39464                                                                        & 15.98                                                                          & \begin{tabular}[c]{@{}c@{}}Learns to deviate\\ from lane and move\\ towards side to \\ avoid collision\end{tabular} \\ \hline
600                                                                  & 295010                                                                      & 49813                                                                        & 9.55                                                                           & Learns to overtake                                                                                                  \\ \hline
\end{tabular}
\caption{\small{Progress of training results with the number of episodes.  The value ``progress" in the table, indicates the reward for lanekeeping behavior which is the forward movement made along the track in one time step. The terms reward and \%colliding steps are same as in table \ref{lane-learning}. Observe how the Progress decreases and then increases. The initial high value is because the agent blindly follows lane, colliding with anyone who comes in between, eventually as the agent tries to navigate safely, collisions decrease as well as the value progress, but the reward is increasing over the training epochs.}}
\label{overtake-training}
\end{center}
\end{table}
\end{itemize}
\begin{table}[H]
\begin{tabular}{|c|c|c|c|}
\hline
\begin{tabular}[c]{@{}c@{}}Number  \\ of Agents\end{tabular} & \begin{tabular}[c]{@{}c@{}}Average total\\ Reward per agent\end{tabular} & \begin{tabular}[c]{@{}c@{}}Average total\\ value of progress\\ per agent\end{tabular} & \begin{tabular}[c]{@{}c@{}}Average\\ Colliding steps \\ per agent\end{tabular} \\ \hline
3                                                            & 56295                                                                   & 8295.2                                                                                & 13                                                                             \\ \hline
5                                                            & 46702                                                                   & 8401.9                                                                                & 10                                                                             \\ \hline
7                                                            & 52798                                                                   & 7769.6                                                                                & 20                                                                             \\ \hline
10                                                           & 62418                                                                   & 8088.3                                                                                & 13.6                                                                           \\ \hline
12                                                           & 6.7859                                                                  & 7909.4                                                                                & 14                                                                             \\ \hline
15                                                           & 76189                                                                   & 7742                                                                                  & 8.7                                                                            \\ \hline
\end{tabular}
\caption{\small{Number of agents indicate the number of agents which were following overtaking behavior. The values, reward and progress are cumulative over all time steps in an episode, averaged over 20 episodes. Average colliding steps are also defined over all time steps in an episode. We observe that the values of progress and collisions have remained in the range 9-14, with an outlier when number of agents was 7. Similarly, progress has also remained in range of 7.7k to 8.2k. This indicates that the performance of agents was not affected by the increase in number of agents, which further justifies the scalability of the architecture. Lastly, the values in reward are increasing, since reward is proportional to total number of agents in the scene (the term, (n-racePos)), which causes this linear increase in average reward values. }}
\label{overtake-testing}
\end{table}
Lastly, we compare PS-DDPG with single agent learned DDPG, in table \ref{overtake-compare}. 
Similar to lanekeeping case, we experimented replicating results for overtaking using multiple DDPG learners. Even though, initial weights of the network were initialized with lanekeeping stable weights, the agents behind the first two agents, could not learn the overtaking behavior. Unfortunately, in multiple DDPGs, even after initialization with lanekeeping rewards, only first two agents were able to learn something. The other agents did not learn anything. The resulting learned behavior of first two agents was equivalent to the single DDPG training behavior.
The number of training episodes required, after curriculum learning, for DDPG were 1k, for PS-DDPG were 600 and for multiple DDPGs together were more than 2k. 
\subsection{Learning cooperative multiple behaviors }
\begin{figure}[H]
\begin{center}
\includegraphics[width=0.45\textwidth]{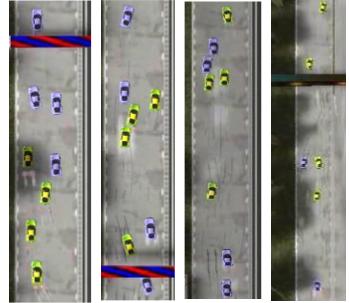}
\caption{\small{Results when multiple behavior were learned using one shared Actor-Critic Network. The yellow agents are overtaking agents and blue agents are lanekeeping agents. Both the behaviors have been learned by a single Actor-Critic instance. The blue cars cooperate with the yellow cars, by slowing down in the start, once the yellow cars have overtook, the blue cars also speed up to increase their own reward which corresponds to the progress along the lane.}}
	\label{multi-behavior}
    \vspace{-2em}
\end{center}
\end{figure}

We learned the two behaviors using a shared network. The reward function have been described in section \ref{implementation}. The training process required required around 1.5k episodes to converge. Our main observations indicated that the lanekeeping agents moved slowly when in presence of other agents, they cannot distinguish the other agents as lanekeeping or overtaking. When they are not in vicinity of other agents, their move with higher velocities. Similarly, overtaking agents are always high sped, they do not compete with other agents, since competition would lead to instability, had the other agent been a overtaking agent. They instead learn how to change lanes smoothly in order to overtake and once they overtake, they maintain the speed to stay ahead in lane. Our quantitative results are shown in figure \ref{multi-behavior} and table \ref{multibehave-test}.
\begin{table}[t]
\begin{center}
\begin{tabular}{|l|c|c|c|}
\hline
\textbf{CaseA}                             & Reward                        & Progress                     & Collisions                   \\ \hline
\multicolumn{1}{|c|}{Overtaking}  & 828.7944                      & 78.7944                      & 23.6667                      \\ \hline
\multicolumn{1}{|c|}{Lanekeeping} & 39.9274                       & 39.9274                      & 55.6667                      \\ \hline
\multicolumn{1}{|c|}{Total}       & 434.3609                      & 59.3609                      & 79.3333                      \\ \hline
\multicolumn{4}{|l|}{\textbf{CaseB}}                                                                                                     \\ \hline
\multicolumn{1}{|c|}{Lanekeeping} & 45.4132                       & 45.4132                      & 87.3333                      \\ \hline
\multicolumn{1}{|c|}{Overtaking}  & 703.1208                      & 64.1625                      & 54.6667                      \\ \hline
Total                             & \multicolumn{1}{l|}{561.4005} & \multicolumn{1}{l|}{82.1817} & \multicolumn{1}{l|}{142}     \\ \hline
\multicolumn{4}{|l|}{\textbf{CaseC}}                                                                                                     \\ \hline
Lanekeeping                       & \multicolumn{1}{l|}{52.8251}  & \multicolumn{1}{l|}{52.8251} & \multicolumn{1}{l|}{46.6667} \\ \hline
Overtaking                        & \multicolumn{1}{l|}{711.8831} & \multicolumn{1}{l|}{67.5081} & \multicolumn{1}{l|}{64.3333} \\ \hline
Total                             & \multicolumn{1}{l|}{382.3541} & \multicolumn{1}{l|}{60.1666} & \multicolumn{1}{l|}{111}     \\ \hline
\end{tabular}
\caption{\small{Analysis of results when the two behaviors were learned simultaneously. Case A refers to the orientation, when all he overtaking agents were placed ahead of all the lanekeeping agents, Case B is vice versa and Case C is all the agents were randomly mixed withe each other. We have used 8 agents, 4 of each type. Here, reward and progress are averaged over timesteps for each agent, over 20 episodes. For all the three orientations the value of reward and progress are lying in similar ranges, indicating the stability of the algorithm across diverse scenes. The collision values are higher than single behavior learning, since the scenes are more complex now.}
\label{multibehave-test}}
\end{center}
\end{table}

\begin{figure}[H]
\includegraphics[width=\linewidth]{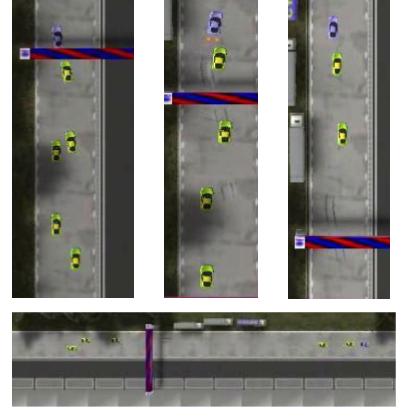}
\caption{\small{Results when traditional DDPG was used for multiple agents. In 6 agent case, the last 3 agents cannot learn to drive. They are stuck at local minima, during training they collided with the front cars and gained negative rewards, henceforth they learned not to move forward.}}
	\label{lanekeeping_ddpg}
\end{figure}
\section{CONCLUSIONS} \label{conclusions}
Parameter Sharing is a well known concept in multi agent systems, we extended it for Deep Deterministic Policy Gradients for a particular case of simulated highway behaviors. The homogeneous nature of the agents, enabled sharing the replay buffer, hence each agent now has a plethora of experiences. The network is updated $N$ times in each time step, because of which the algorithm converges faster. In the cases when connecting additional agents does not require heavy resources, PS-DDPG can be used to speed up the training and to learn more generically. 
Apart from its advantages over DDPG, it serves as a fast-asynchronous multi agent learning algorithm. With a correct formulation of reward function and state vector, multiple behaviors can be learned jointly, an example of which is shown in this work. Another advantage the current work offers is scalability. This can be used to generate behavioral traffic in simulations. Given the interests in autonomous driving, a simulator which provides scalable traffic will help accelerate many complex research statements.





\bibliographystyle{unsrt}
\bibliography{root}

\end{document}